# Credentials in the Occupation Ontology

John Beverley[1,2,3*], Robin McGill[4*], Sam Smith[5], Jie Zheng[5], Giacomo De Colle[1,2], Finn Wilson[1,2], Matthew A. Diller[6], William D. Duncan[6], William R. Hogan[7], and Yongqun He[5]

1  University at Buffalo, NY, USA.
2  National Center for Ontological Research, NY, USA.
3  Institute for Artificial Intelligence and Data Science, NY, USA.
4  Alabama Commission on Higher Education, AL, USA.
5  University of Michigan Medical School, Ann Arbor, MI, USA.
6  University of Florida, Gainesville, FL, USA.
7  Medical College of Wisconsin, Milwaukee, WI, USA.

* contributed equally.

#### Abstract

The term "credential" encompasses educational certificates, degrees, certifications, and government-issued licenses. An occupational credential is a verification of an individual's qualification or competence issued by a third party with relevant authority. Job seekers often leverage such credentials as evidence that desired qualifications are satisfied by their holders. Many U.S. education and workforce development organizations have recognized the importance of credentials for employment and the challenges of understanding the value of credentials. In this study, we identified and ontologically defined credential and credential-related terms at the textual and semantic levels based on the Occupation Ontology (OccO), a BFO-based ontology. Different credential types and their authorization logic are modeled. We additionally defined a high-level hierarchy of credential related terms and relations among many terms, which were initiated in concert with the Alabama Talent Triad (ATT) program, which aims to connect learners, earners, employers and education/training providers through credentials and skills. To our knowledge, our research provides for the first time systematic ontological modeling of the important domain of credentials and related contents, supporting enhanced credential data and knowledge integration in the future.

## 1. Introduction

Broadly speaking, *credentials* are "status indicators" that make possible new opportunities for their holders [1, 2]. For example, a passport is a credential that signifies citizenship while permitting its holder to travel internationally; a parking pass is a credential that signifies its holder is among those permitted to leave a vehicle unattended in a specific area. In education and employment contexts, credentials are often designed to signify their holder has certain skills, abilities, and knowledge that are valuable within a given professional context. Job seekers often leverage *occupational credentials* – such as diplomas, certificates, licensures, etc. – as evidence that desired qualifications are satisfied by their holder. For example, an academic degree may signify its holder has completed a curriculum that involves the attainment of skills and knowledge, which in turn permits the *credential holder* to be considered as a plausible candidate for job postings requiring such skills and knowledge.

As credentials have become a kind of "currency" in the labor market [3], the number of credentials has continued to grow. In 2022, over 1 million unique employment credentials were identified in the United States [4], half of which emerged from non-academic providers, reflecting the rise of non-traditional education pathways among job-seeking adults [5]. The sheer number of available credentials, however, threatens to undermine their utility for both hiring agencies and job seekers [6].

Job seekers rely on the information provided in job descriptions to deduce the desired qualifications sought by agencies, while hiring agencies must sift through resumes to identify candidates whose education and work experience align with those descriptions. For every trustworthy credential that clearly and accurately signifies the competency of its holder, there are surplus credentials of questionable quality, and few, if any, hiring managers have the resources needed to verify credential quality amidst so many options. This challenge is exacerbated further when observing that hiring agencies must often compare credentials that are described differently by distinct providers often using different languages, even within the same discipline [7].

In the United States, education and workforce development policy organizations have recognized both the importance of credentials for employment and challenges of understanding the value of credentials. A theme among proposals for addressing these challenges is the establishment of standards for representing, storing, and displaying information pertaining to credentials, especially "non-degree credentials" [8, 9], which have significant variation in terms of quality and value in the labor market. To ensure that public workforce development funding is spent on high-quality training programs, several states have taken up "credential transparency" as an important policy issue [10]. For instance, the Washington Workforce Training and Education Coordinating Board coordinated the registration of credentials in the state into a linked data platform [11]. The New England Board of Higher Education established a minimum data policy for the six states in its region establishing reporting standards for important features of credentials, such as duration and cost [12]. One of the most comprehensive efforts to collect, evaluate, and publish credential information is underway in Alabama. Since 2021, a broad coalition of industry experts and partners from state education and workforce agencies has worked to create a statewide registry of credentials that can help job seekers gain the skills needed to fill in-demand jobs [13].

These examples have in common goals of creating common reporting and annotation standards for credential data. Each organization above uses ontology artifacts maintained by the non-profit organization Credential Engine, which aims to create a national registry of credentials, tagged with common metadata, that will enable users to search for and differentiate among the myriad credentials currently offered by education and training providers [12]. Credential Engine has developed the Credential Transparency Description Language (CTDL), a family of standards for making descriptions of job-related credentials and other available resources [14]. As a description language, its main purpose is to support data description, data exchange, and schema generation. CTDL boasts an extensive collection of credential-related resources developed over several years through working groups with stakeholders, and as a result, it has been widely adopted to support tools that can more easily compare the salient features of credentials and help eliminate confusion within the credential marketplace.

Nonetheless, there are opportunities to refine the ontological underpinnings of CTDL so that it can better integrate with data outside its system, specifically data related to occupations, which is essential for understanding the value of credentials. This paper explores the potential of the Occupational Ontology (OccO) to model relationships between credentials and other occupation-related terminology, which the authors believe will be of great use to those working to align education and workforce development systems. OccO [15] was initiated in the interest of providing a semantic layer connecting occupation coding standards, such as those found in the US Bureau of Labor Statistics Standard Occupational Classification (US SOC) [16]. OccO adopts the Basic Formal Ontology (BFO) as a top-level ontology architecture [17] and therefore contributes to the extensive ecosystem of BFO-based ontologies, which represent content common to all areas of scientific investigation. In this paper, we report our OccO-based ontological modeling of occupational credentials and semantic relations among them. In addition, we explore the Alabama Talent Triad (ATT) initiative as a use case for credential-related ontology research. ATT is designed to connect credential data with occupational information and individuals' professional history to help job-seekers find educational programs and employment opportunities that align with their professional experiences and aspirations [18]. ATT is thus a particularly fitting use case for ontological modeling at the intersection of credentials and occupations.

## 2. Occupational Credentials

We start by considering the term "credential." Stemming from the Latinate root *cred-*, which means "believe," a credential can be understood as a document intended to give rise to a belief about some quality of or distinction concerning a credential bearer. Ontologically, credentials plausibly fall under the Information Artifact Ontology (IAO) class *document*. Credentials have hallmarks that separate them from other sorts of documents.[1] For example, credentials can be broadly distinguished along three axes [19]: those that bear text, e.g. passports, and those that do not, e.g. military insignia; those that display identifying information about the bearer, e.g. credit card, and those that do not, e.g. hotel room keycard; those that can be transferred, e.g. bus ticket, and those that cannot, e.g. diploma. Within this broad framing, credentials in the sense we are concerned with typically bear text, display information about the holder, and cannot be transferred. Another characteristic of credentials is that they must be in some sense within a holder's control. For example, driving without one's license will earn you a sanction. Similarly, college graduates must often have physical or digital copies of diplomas sent to hiring agencies to verify education.

Occupation-relevant credentials often inherit many of these general features of credentials, as evidenced by existing definitions of such credentials. For example, according to the W3C recommended Verifiable Credentials Data Model a credential is a set of one or more claims made by an issuer [20]. Similarly, according to the Association for Career and Technical Education (ACTE), education or work-related credentials are "a verification of an individual's qualification or competence issued by a third party with the relevant authority." [21] The U.S. Bureau of Labor and Statistics maintains that "Credentials include training time required as a condition of hiring, which often results in certifications, licenses, or educational certificates and are part of the education, training, and experience requirements." [22] CTDL defines a credential as a "Qualification, achievement, personal or organizational quality, or aspect of an identity typically used to indicate suitability" and requires that a credential be associated with an organization that offers or issues or recognizes it.

Synthesizing these definitions of "credential" within an ontologically precise characterization resulted in the set of definitions listed in **Table 1**, which have been added to OccO. Within OccO's model, credentials are defined as "documents issued by some third party that has the authority to grant the credential." As illustrated in this definition, an ontological characterization of credentials involves reference to a variety of other ontology elements, such as the third party bearing relevant authority (i.e. *credential granting agency*) and *competence, skills, abilities, etc.* **Table 1** lists many such definitions that have been added to OccO in the interest of adequately defining credential and related terms. These definitions in turn provide the foundations for further credential-related definitions as seen below.

**Table 2.** *Key OccO textual definitions of credential and related terms*

| Label | Definition |
|---|---|
| *credential* | A document issued by a third party recognized as bearing the relevant authority to do so, that is designed to be about an entity's competence, qualifications, or authority. |
| *occupational credential* | A credential that is designed to be about an organism's[2] competence or qualifications with respect to an occupation. |
| *occupational credential holder* | An organism that an occupation credential is about. |
| *competence* | A disposition borne by an organism in virtue of training such that, if realized, is realized in the successful performance of a skilled task for which that training was pursued. |
| *credential grantor role* | A role borne by an organization that has been accredited by a quality assurance group through a deontic declaration having action regulation output, which authorizes the organization to bestow credentials according to quality control standards established or enforced by the accrediting group. |
| *credential granting agency* | An organization that bears a credential grantor role. |

---

[1] Credentials are status-indicators but not *status-makers*. Though we must often display credentials to enjoy signified rights, the credential itself does not bestow the right. Rather, credentials are often proxies for education, rights, or honors. Compare the CTDL definition of *license* as a credential that "constitutes legal authority to do…" A license signifies, but does not constitute, legal authority.

[2] Non-human animals may bear credentials. Military dogs, for example, enter duty as non-commissioned officers.

| | |
|---|---|
| *employer role* | A role in human social processes that is realized when the bearer provides a wage or salary in exchange for some labour or services as specified by some declaration.[3] |
| *certificate* | A credential awarded by a credential granting agency designed to signify participation in or completion of a training program requiring the demonstration of some specified knowledge or skill. |
| *certification* | A credential awarded by a professional organization designed to signify the holder's competence as measured against a specified professional benchmark. |
| *academic degree* | A credential awarded by an educational institution designed to signify the satisfaction of requirements established by and under the monitoring of that institution.[4] |
| *license* | A credential awarded by a government or government-authorized agency designed to signify legal authority to engage in specified actions without being liable to sanctions or penalties that would otherwise be associated with those actions. |
| *credential issuing process* | A social act[5] in which a credential granting agency asserts that an organism has satisfied the requirements for being awarded a credential, having as output a credential about the organism. |
| *quality assurance group* | An organization established to set standards of activity by other organizations, evaluate the extent to which organizations satisfy or fail to satisfy those standards, and bestow or revoke the permissions afforded to organizations based on such evaluations. |
| *accredited by* | A relation that holds between a credential granting agency and a quality assurance group when the credential granting agency satisfies accreditation standards set by, and confirmed to have been satisfied by, the quality assurance group. |

Building from *credential* our key design pattern relates instances of *credential* to competencies and the trainees who bear them. A *valid credential* is a credential that accurately signifies the status and permissions of its holder. Hiring agencies have a vested interest in being able to distinguish valid from invalid credentials. Examples of invalid credentials include a forged diploma, certificate offered by degree mill [25], or a degree granted by an agency lacking the authority to do so. Typically diplomas are understood to reflect the completion of a curriculum, passing examinations, maintaining grades above a certain threshold, and so on. Forged diplomas inaccurately signify their holder has completed such training. Similarly, degrees acquired through organizations offering curriculum content falling short of acceptable college standards inaccurately signify holders have completed such training. And if an organization purports to have the authority to grant degrees, but in fact does not, such a degree will inaccurately signify the status of its holder as having a degree granted based on that authority. To represent such relationships, OccO leverages IAO's *is about*, a primitive relation connecting information to entities it is about, which in this case connects instances of *credential* their holders. When credentials are valid, we say, moreover, that the credential is about competencies born by the trainee. **Figure 1** shows how we represent valid credentials. When dealing with invalid credentials, a given credential may be about some credential holder who lacks the advertised competence.

---

[3]Reused from the Ontology for Modeling and Representation of Social Entities (OMRSE) [23].
[4]Most academic degrees are designed to signify the completion of a course of study; honorary degrees are an exception. Nonetheless, honorary degrees signify the satisfaction of requirements established by an educational institution through a waiver of those requirements.
[5]A process that is carried out by a self-conscious being and is spontaneous, directed towards another conscious being, and needs to be perceived. [24].

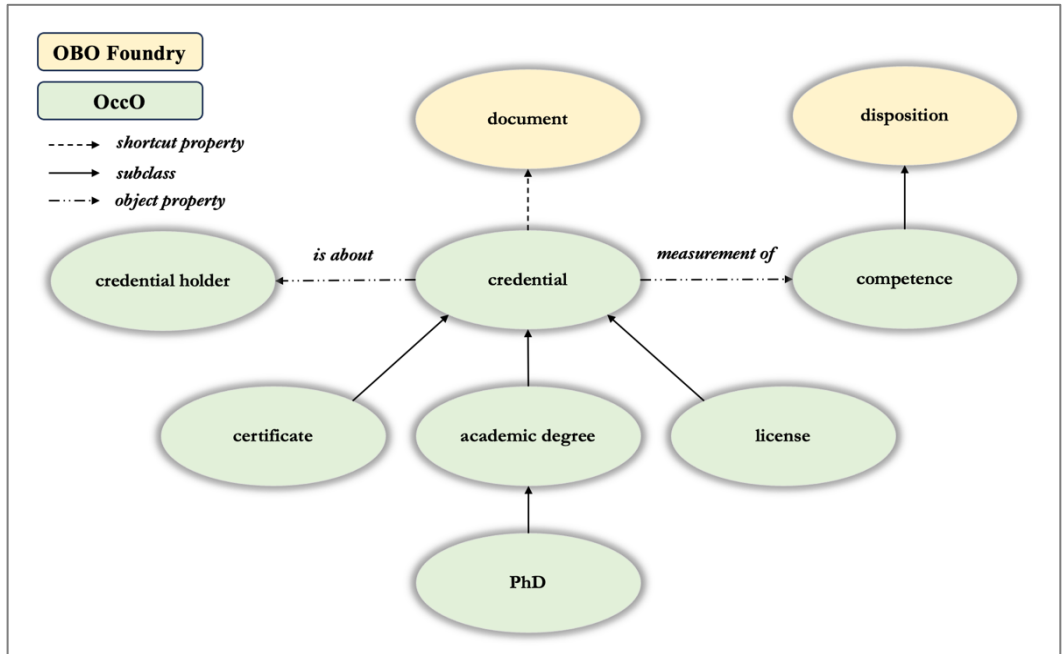

**Figure 1**: *Key OccO Credential Classes*

**Figure 2** illustrates a scenario in which a valid credential has been granted by an organization with the appropriate authority. A credential granting agency makes a deontic declaration which produces an action regulation that prescribes the role of this agency. A deontic declaration is a social act that creates or revokes a deontic role, which is a role regarding how a person should behave [26]. This declaration is realized by and produces an action regulation, which is a directive information content entity that prescribes an act as required, prohibited, or permitted. To reflect the granting of an invalid credential, we need only break the recognition of the *credential granting agency* by the *quality assurance group*.

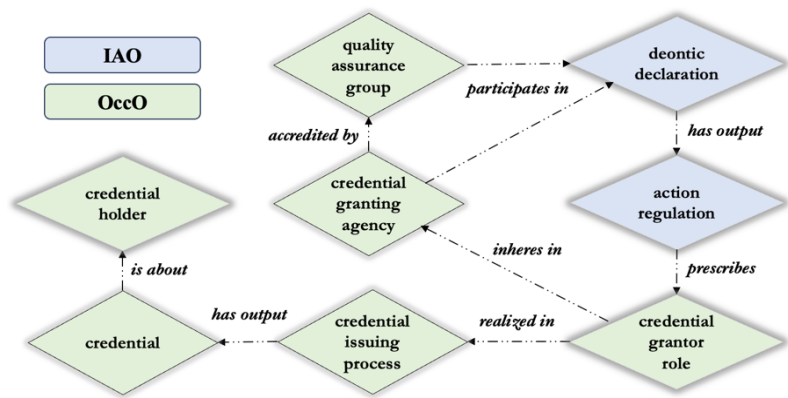

**Figure 2**: OccO *Relations of Deontic Authority*

## 3. OccO and the Alabama Talent Triad

The expansion of OccO to model relationships among credential-related terms will be valuable for policymakers and business leaders seeking to align education/training and workforce development systems. OccO developers have been working with one such example in the Alabama Talent Triad (ATT), a "comprehensive skills-based talent marketplace" [18], led by the Alabama Governor's Office of Education and Workforce Transformation in partnership with the Alabama Commission on Higher Education and many other public and private entities. The ATT program serves as an important use case and key motivation for our OccO credential-related ontological modeling.

Launched in Fall 2023, the ATT integrates three digital applications, each designed for a different audience of users [18]:

1) The Digital Wallet is designed to be used by *credential holders* as a verified record of their educational and employment experience.
2) The Credential Registry enables *credential granting agencies* to describe the credentials that they offer, with a particular focus on those credentials that align with Alabama's Statewide In-Demand Occupations List and meet criteria for quality and transparency.
3) The Skills-Based Job Description Generator enables *employers* to create customized job descriptions for recruitment and hiring that are organized around competencies.

Each component relies on different kinds of data. The Digital Wallet is organized around the principles of Learning and Employment Records (LERs) [1, 27, 28], which include personally identifiable information, such as an individual's education records, and therefore must adhere to certain data privacy and security standards. The Credential Registry is organized around credential information and uses elements of the Credential Transparency Description Language (CTDL), developed and maintained by Credential Engine [7]. The Skills-Based Job Description Generator is perhaps the most unique of the ATT three applications. It relies on a growing set of occupation-based competency statements that were initially developed by the Alabama Committee on Credentialing and Career Pathways (ACCCP) and its industry-based Technical Advisory Committees [29]. Through the work of EBSCOed [30] and other partners, the competency statements have been standardized and tagged so that they can be deployed as templates for competency-based job descriptions, which employers can then customize for their specific context.

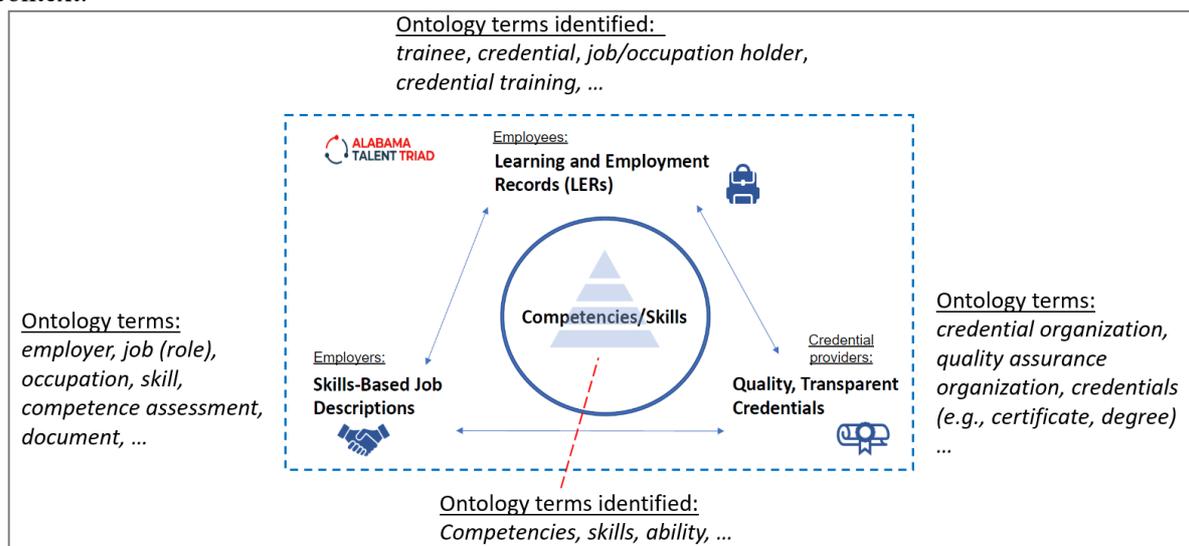

**Figure 3**: *Alabama Talent Triad (ATT) Design and ontology modeling.*[6]

As represented in **Figure 3**, the functionality of the Talent Triad is based on the premise that information about competencies or skills can be represented within all three data sets, and ATT's developers refer to this shared set of competency/skills information as the "Alabama Competency Ontology" or "Occupational DNA" [31]. The vision of ATT is that automated methods will be developed to analyze competency information in the interest of making individualized predictions or "matches" for each user. The result would be credential holders receiving recommendations for job openings that match their skills, as well as for further training opportunities to enhance their job prospects. In addition, credential providers would receive recruitment suggestions for individuals who may be likely to enroll in their offerings based on their competencies and goals, while employers would receive recommendations about candidates likely to have the skills required to fill an open position. With its focus on competencies and skills, the ATT fits a broader trend across the United States towards the adoption of "skills-" or "competency-based" occupation education models [32, 33, 34], which emphasize skills and competencies as the primary consideration for hiring, rather than degrees earned [35].

---
[6]The original ATT design is laid out inside the box while OccO terms inhabit the outer layer and relate to parts of the ATT model.

Accurately connecting credentials to expected competencies requires careful, detailed, representations of credentials, competencies, and possible pathways. Since 2022, the Alabama credential team (including Dr. Robin McGill, a co-author of this paper), has been collaborating with OccO developers with the goal to use the OccO to ontologize ATT content. **Figure 4** illustrates one of the design patterns emerging from our collaboration. We import the class *human* from the NCBITaxon [36] as a parent to those who hold occupations. An *occupation holder* is someone bearing either an occupation role or an occupation disposition. This class captures the bearers of occupations, such as a pharmacist, welder, ontologist, etc.

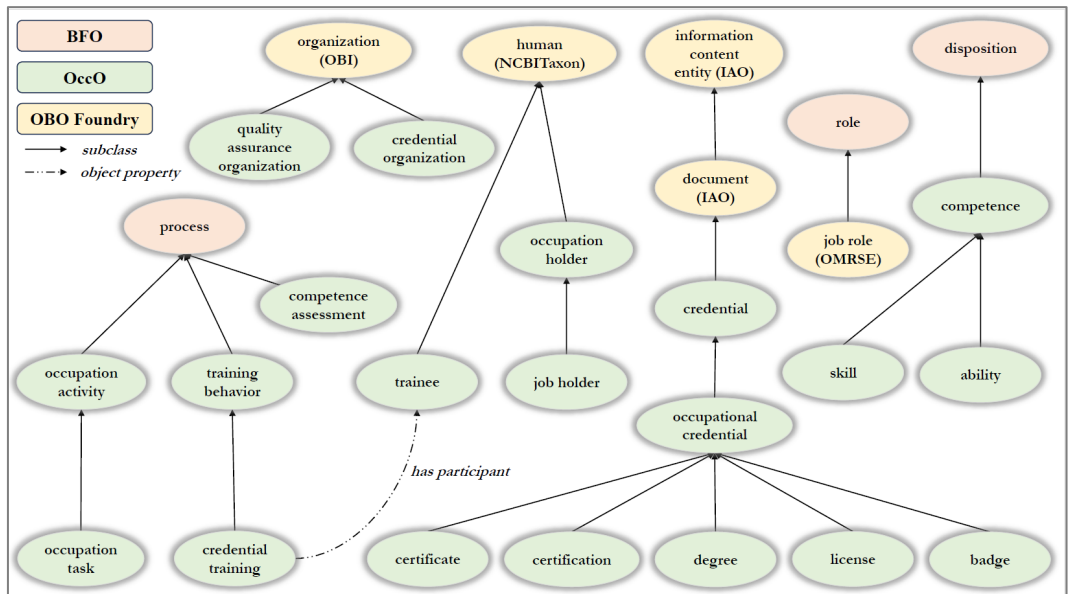

**Figure 4**: *Key OccO Credential Classes*

Within ATT, the term "skill" is used interchangeably with "competency," most often in plural form, i.e. "skills" and "competencies." In OccO (**Figure 4**), we classify both *skill* and *ability* under *competence* (used synonymously with "competency"), which is defined as a subclass of *disposition*. OccO currently includes 35 skill types and 52 ability types, primarily derived from the US SOC and O*NET [37] classifications. We also noted that ATT and CTDL have also defined many skills and abilities, which invite future comparison and harmonization.

In addition to the identification and definition of terms related to ATT, we have also used the OccO to develop semantic relationships among different ATT terms, which are illustrated in **Figure 5**. For example, this design shows that *trainees* participate in *trainee activities* (behaviors), and in specific cases, credential trainees participate in *credential training*. The objective of credential training is to obtain credentials. However, the relation *planned objective* appearing in **Figure 5** appears problematic as a relation between *credential training* and *credential*, since credential training need not always have as its objective the obtaining of a credential. We leverage the relation *evidence of* to show that a credential is evidence of some *competence*, i.e. it is designed to signify that the holder has specific competence. Furthermore, our model allows us to represent assessments of competence, where competence is a qualification for *occupation activity*, which then has participant of *occupation holder*. Making our way around to the individual credential holder aiming to obtain employment, we say a job holder is an *occupation holder* who works for an *employer*. Here we define employer as a subclass of organization. Note that trainees can be existing employees who are engaged in additional training or job-seekers who do not have a job yet. The ontology model proposed here allows for seamless and logical links among the three important groups of the ATT program: trainees (job-seekers), employers, and credential providers.

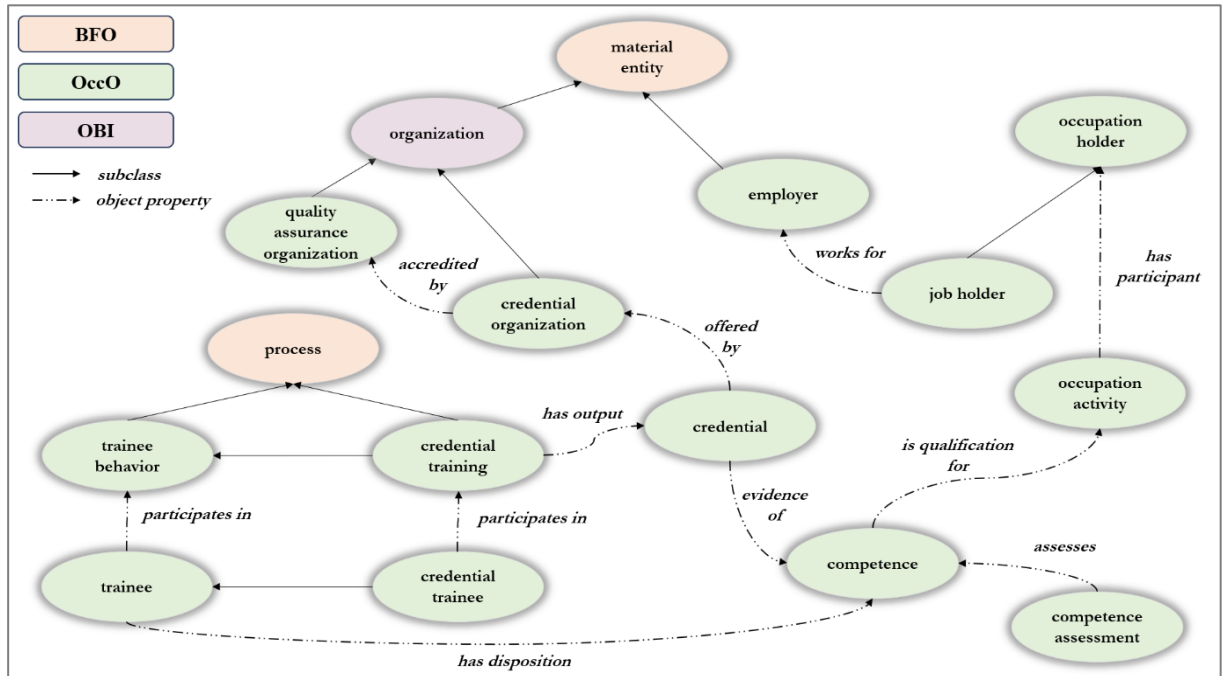

**Figure 5.** *Relations among ATT-Related Credential Entities*

## 4. Discussion

Several contributions are reported in this article. First, we introduced ontological definitions of credential and credential-related terms. Second, we provided ontological modeling of credential and its authorization. Third, we developed a top-level OccO hierarchy of credential related terms and a design pattern, which were further used to evaluate and support the Alabama Talent Triad (ATT) program. To our knowledge, our research provides for the first time the systematic ontological modeling of the important domain of credentials and related contents, supporting enhanced credential data and knowledge integration in the future. OccO is being developed to address both existing and novel occupations, and with this report, we describe the incorporation of occupational credentials into OccO. This combination of credentials and occupations into a uniform ontological framework will be the first systematic ontological characterization of these two fields.

There is a lot of potential for the ontological frameworks put forward here to help promote semantic interoperability between distinct but overlapping communities related to credentials. Specifically, we have noted the opportunity to work more closely with the developers of CTDL in the interest of aligning efforts. Likewise, the OccO developers have recently become aware of systematic efforts to establish data standards for verifiable credentials that have been recognized by the W3C [20], and we hope to collaborate with that working group around interoperability. We also look forward to continued collaboration between our OccO development team and the ATT program. Our ontological modeling of the core ATT design provides the first step for further collaboration in the future. We believe our ontological modeling will significantly lift the ATT program to make it more interoperable and recognizable as an emerging best practice in skills-based talent development efforts. We believe our ontological modeling of occupational credentials will benefit ATT by validating the approach it has taken to articulating and developing credentials within the state. On the other hand, ATT will benefit OccO by incorporating the important characteristics of occupational credentials.

## 5. Acknowledgements

Many thanks to Gianluca Bortoletto and Jan Luts of the European classification of Skills, Competences, and Occupations (ESCO) team for discussions on previous versions of OccO, as well as Nick Moore, director of the Alabama Governor's Office of Education and Workforce Transformation, for helpful discussions.